\documentclass[sigconf]{acmart}
\AtBeginDocument{%
  }

\usepackage{subcaption}
\usepackage{multirow}
\usepackage{url}
\newcommand{\ours}{UNIStainNet}
\newcommand{\first}[1]{\textbf{#1}}
\newcommand{\secondbest}[1]{\underline{#1}}

\renewcommand\footnotetextcopyrightpermission[1]{}
\setcopyright{none}
\acmConference{Preprint}{}{}{}
\acmDOI{}

\begin{document}

\title{UNIStainNet: Foundation-Model-Guided Virtual Staining of H\&E to IHC}

%% ── Authors ─────────────────────────────────────────────────────
%% TODO: Replace the block below with your real author information.
%% Format: \author{} \affiliation{} \email{} per author.
%% Example:
%%
%% \author{Saurav Lastname}
%% \affiliation{
%%   \institution{Your Institution}
%%   \city{City}
%%   \country{Country}
%% }
%% \email{saurav@example.edu}
%%
\author{Jillur Rahman Saurav}
\email{mxs2361@mavs.uta.edu}
\affiliation{%
  \institution{Department of Computer Science, University of Texas at Arlington}
  \city{Arlington}
  \state{Texas}
  \country{USA}
}

\author{Thuong Le Hoai Pham}
\email{tlp5359@mavs.uta.edu}
\affiliation{%
  \institution{Department of Computer Science, University of Texas at Arlington}
  \city{Arlington}
  \state{Texas}
  \country{USA}
}

% \author{Manfred Huber}
% \email{huber@cse.uta.edu}
% \affiliation{%
%   \institution{Department of Computer Science, University of Texas at Arlington}
%   \city{Arlington}
%   \state{Texas}
%   \country{USA}
% }

\author{Pritam Mukherjee}
\email{pritam.mukherjee@stjude.org}
\affiliation{%
  \institution{Department of Radiology, St. Jude Children's Research Hospital}
  \city{Memphis}
  \state{Tennessee}
  \country{USA}
}

\author{Paul Yi}
\email{paul.yi@stjude.org}
\affiliation{%
  \institution{Department of Radiology, St. Jude Children's Research Hospital}
  \city{Memphis}
  \state{Tennessee}
  \country{USA}
}

\author{Brent A. Orr}
\email{brent.orr@stjude.org}
\affiliation{%
  \institution{Department of Pathology, St. Jude Children's Research Hospital}
  \city{Memphis}
  \state{Tennessee}
  \country{USA}
}

\author{Jacob M. Luber}
\email{jacob.luber@stjude.org}
\affiliation{%
  \institution{Department of Radiology, St. Jude Children's Research Hospital}
  \city{Memphis}
  \state{Tennessee}
  \country{USA}
}  %% <-- ensure this file contains real (non-anonymized) author info

\renewcommand{\shortauthors}{Saurav et al.}

\begin{abstract}

Virtual immunohistochemistry (IHC) staining from hematoxylin and eosin (H\&E) images can accelerate diagnostics by providing preliminary molecular insight directly from routine sections, reducing the need for repeat sectioning when tissue is limited.
Existing methods improve realism through contrastive objectives, prototype matching, or domain alignment, yet the generator itself receives no direct guidance from pathology foundation models.

We present \ours{}, a SPADE-UNet conditioned on dense spatial tokens from a frozen pathology foundation model (UNI), providing tissue-level semantic guidance for stain translation.
A misalignment-aware loss suite preserves stain quantification accuracy, and learned stain embeddings enable a single model to serve multiple IHC markers simultaneously.

On MIST, \ours{} achieves state-of-the-art distributional metrics on all four stains (HER2, Ki67, ER, PR) from a single unified model, where prior methods typically train separate per-stain models.
On BCI, it also achieves the best distributional metrics.
A tissue-type stratified failure analysis reveals that remaining errors are systematic, concentrating in non-tumor tissue.
Code is available at \textbf{\url{https://github.com/facevoid/UNIStainNet}}.

\end{abstract}


\maketitle

\section{Introduction}
\label{sec:intro}

Immunohistochemistry (IHC) staining is central to molecular profiling in cancer diagnostics, guiding treatment decisions for biomarkers such as HER2, Ki67, ER, and PR in breast cancer~\cite{wolff2018her2}.
However, IHC requires dedicated tissue sections, specialized reagents, and multi-day turnaround, and these resources are limited in many clinical settings and constrained by available tissue.
Virtual staining, computationally generating IHC images from routinely available H\&E slides, offers a complementary tool that can aid pathologists by providing rapid preliminary assessments, reducing tissue consumption, and enabling molecular screening where IHC infrastructure is unavailable~\cite{liu2022bci,liu2024pspstain,li2023odagan}.
Beyond direct clinical use, virtually generated IHC images can also serve as input to computer-aided diagnosis systems for automated biomarker scoring and quantification.

A central difficulty in training virtual staining models is the lack of pixel-aligned ground truth.
In standard clinical practice, H\&E and IHC are performed on separate consecutive tissue sections that are stained and scanned independently.
This introduces 10--50\,px misalignment from morphological variation between cuts, tissue deformation during processing, and registration error, making pixel-level losses unreliable and complicating evaluation with metrics such as SSIM and PSNR~\cite{li2023asp,klockner2025h}.

Existing methods have advanced generation quality through careful loss design and domain-specific feature engineering to handle these challenges~\cite{li2023asp,liu2024pspstain,li2023odagan}.
Contrastive objectives down-weight inconsistent regions~\cite{li2023asp}, prototype matching avoids spatial correspondence~\cite{liu2024pspstain}, and domain alignment strategies reduce reliance on pixel-level supervision~\cite{li2023odagan}.
However, leveraging pathology foundation models as a direct conditioning signal for the generator remains underexplored: one study~\cite{li2023odagan} derives segmentation masks from UNI~\cite{chen2024uni} embeddings, but the generator itself receives no foundation-model signal.
Such models, trained on large-scale histopathology data, learn tissue representations that have proven effective across diverse pathology tasks~\cite{chen2024uni}, making them a natural source of conditioning information for stain translation.
To explore this direction, we present \ours{} with the following contributions:

\begin{enumerate}
    \item \textbf{Dense UNI spatial conditioning.}
    A SPADE-UNet~\cite{park2019spade} generator conditioned on spatial tokens from a frozen UNI~\cite{chen2024uni} backbone via SPADE modulation at each decoder stage.

    \item \textbf{Misalignment-aware loss design.}
    A loss suite designed to avoid pixel-level supervision:
    perceptual losses at reduced resolution,
    an unconditional discriminator,
    feature matching for texture statistics,
    and a DAB intensity loss for stain quantification.

    \item \textbf{Unified multi-stain generation.}
    Learned stain embeddings via FiLM modulation enable a single model to serve all four IHC markers.

    \item \textbf{Tissue-type stratified failure analysis.}
    We provide the first systematic characterization of \emph{where} generation fails, using zero-shot tissue classification to stratify errors by morphological context across both datasets.
\end{enumerate}

We validate against leading published methods on two public benchmarks at the standard $512\!\times\!512$ crop protocol, achieving state-of-the-art distributional metrics on both MIST~\cite{li2023asp} and BCI~\cite{liu2022bci}.
The architecture extends to native $1024\!\times\!1024$ generation with minimal parameter overhead, improving stain accuracy on both datasets.

% Architecture figure — placed early so readers see the method overview
\begin{figure*}[t]
\centering
\includegraphics[width=\textwidth]{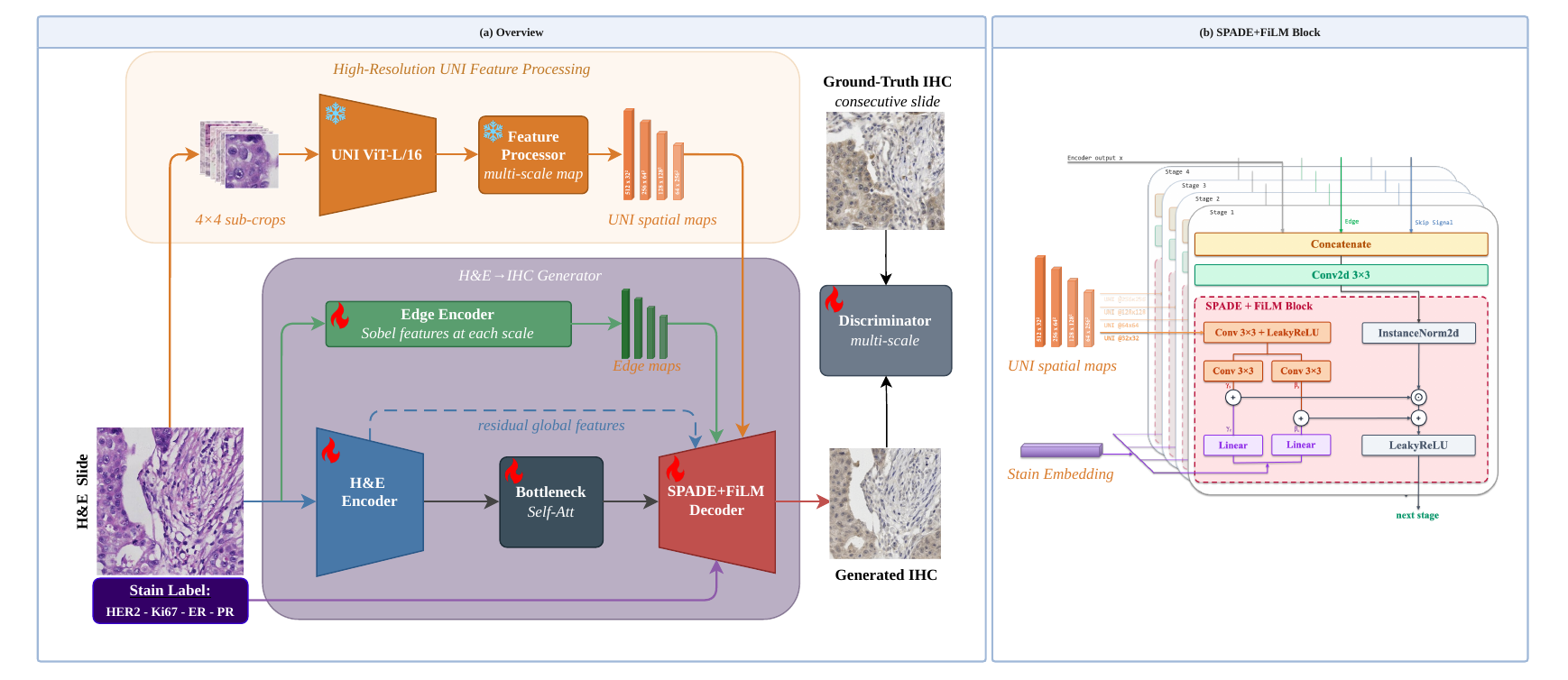}
\caption{\textbf{\ours{} architecture.} \textbf{(a)} Overview: The H\&E image is split into $4\!\times\!4$ sub-crops and processed by a frozen UNI ViT-L/16 to produce multi-scale spatial maps. A CNN encoder compresses the H\&E input through a self-attention bottleneck; the SPADE+FiLM decoder then receives UNI spatial maps via SPADE modulation, edge features via concatenation, and a stain embedding via FiLM, with encoder skip connections preserving fine-grained detail. An unconditional multi-scale discriminator provides adversarial training. \textbf{(b)} SPADE+FiLM block detail: at each decoder stage, edge and skip features are concatenated and convolved, then modulated by UNI spatial maps (spatially-varying $\gamma_s, \beta_s$) and the stain embedding (channel-wise $\gamma_c, \beta_c$), combined additively after instance normalization.}
\label{fig:architecture}
\end{figure*}

\section{Related Work}
\label{sec:related}

\subsection{H\&E-to-IHC Virtual Staining}

Virtual staining computationally generates IHC images from routinely available H\&E slides, reducing the cost and turnaround of chemical staining.
Early approaches rely on paired pixel-level losses: the BCI benchmark introduced Pyramid Pix2Pix~\cite{liu2022bci}, which trains a multi-scale encoder-decoder with Gaussian-pyramid reconstruction losses.
However, pixel-level supervision is degraded by the spatial misalignment inherent in consecutive-section pairs, biasing generators toward blurry, spatially averaged outputs~\cite{klockner2025h}.

Subsequent GAN-based methods can be grouped by how they handle the misalignment challenge and inject domain knowledge.
\emph{Contrastive approaches} relax the need for pixel-aligned supervision: CUT~\cite{park2020cut} introduces PatchNCE for patch-level correspondence, ASP~\cite{li2023asp} adds similarity-based down-weighting of inconsistent regions (and contributes the MIST multi-stain dataset), and ODA-GAN~\cite{li2023odagan} disentangles expression- and morphology-dependent features via orthogonal projection with multi-layer MMD alignment.
\emph{Domain-knowledge approaches} incorporate stain-specific priors: PSPStain~\cite{liu2024pspstain} uses Beer--Lambert-derived Focal Optical Density maps with prototype matching, TDKStain~\cite{peng2024advancing} integrates HER2-specific scoring criteria, and TA-GAN~\cite{jiang2026topology} enforces topology-aware consistency via graph neural networks.
\emph{Optimal-transport approaches} such as SIM-GAN~\cite{guan2025supervised} mine supervised correspondences via Wasserstein-distance feature matching, and its successor USI-GAN~\cite{peng2025usigan} extends this to unbalanced self-information transport across the domain gap.
These methods progressively layer multi-stage feature engineering on top of a shared CUT-derived ResNet generator.
\ours{} takes a different route: rather than engineering supervision signals, we condition the generator directly on dense spatial tokens from a frozen pathology foundation model.

Diffusion-based approaches have recently been applied to this task.
PASB~\cite{pasb} formulates the translation as a Schr\"odinger Bridge optimal-transport problem and guides generation through an IRS-derived classification constraint and a reference-similarity refinement step at each denoising iteration.
Star-Diff~\cite{liu2025stardiff} frames virtual staining as a restoration problem, pairing a deterministic residual network with a stochastic denoising network for structural fidelity.
Diffusion-based approaches involve iterative inference, whereas GAN-based methods (including \ours{}) require only one pass.

\subsection{Foundation Models in Computational Pathology}

Central to our approach is the use of pathology foundation models as a conditioning signal.
These models, large vision transformers pretrained on millions of histology images, have demonstrated strong transfer across diverse downstream tasks.
UNI~\cite{chen2024uni}, pretrained on over 100 million histopathology images, produces 1024-dimensional patch tokens that encode tissue morphology.

In the virtual staining literature, pathology foundation features have so far been used only for auxiliary purposes.
ODA-GAN~\cite{li2023odagan} feeds UNI embeddings through ABMIL and Grad-CAM to produce spatial masks that partition contrastive samples, but these features remain external to the generator.
PASB~\cite{pasb} derives a classification-based pathological constraint from adjacent IHC sections, independent of the generative backbone.

\ours{} is, to our knowledge, the first H\&E-to-IHC method to condition the generator on dense spatial foundation model features, giving it access to tissue-level semantic context unavailable to methods that rely solely on pixel-space encoder features (Section~\ref{sec:uni_features}).

\subsection{Handling Consecutive-Section Misalignment}

Datasets pairing H\&E and IHC from consecutive tissue sections (BCI~\cite{liu2022bci}, MIST~\cite{li2023asp}) exhibit inherent spatial misalignment because the two stains are applied to adjacent, not identical, slices.
Na\"ive application of pixel-level paired losses on such data causes the generator to absorb misalignment as part of the target distribution, producing spatially inconsistent outputs.

Existing approaches mitigate this in several ways.
ODA-GAN~\cite{li2023odagan} reduces reliance on pixel correspondence through a contrastive-adversarial formulation with orthogonal feature decoupling, at the cost of discarding much of the paired supervision signal.
ASP~\cite{li2023asp} measures per-patch agreement between generated and ground-truth embeddings and progressively suppresses the loss at locations where correspondence is low.
PSPStain~\cite{liu2024pspstain} replaces pixel-level alignment with cross-image prototype matching, comparing aggregated tumor representations that are invariant to spatial shifts.

\ours{} takes a different approach: every loss component is designed to tolerate misalignment while retaining paired supervision.
Perceptual losses operate at resolutions (128, 256\,px) where the misalignment becomes sub-pixel; L1 is computed at 64\,px.
The discriminator is unconditional, since conditioning on misaligned pairs would cause it to learn spatial inconsistency as ``real.''
Structural supervision (edge loss) acts on the pixel-aligned H\&E$\to$generated axis rather than the misaligned H\&E$\to$IHC axis.
This design preserves the benefits of paired training while avoiding the artifacts introduced by na\"ive pixel-level supervision.

\section{Methodology}
\label{sec:method}

\subsection{Overview}

The generator translates an H\&E input into an IHC output in a single forward pass, guided by three signals: dense spatial tokens from a pretrained pathology model, structural edge maps, and a stain identity embedding.
Formally, given an H\&E image $\mathbf{x}_{\text{HE}} \in \mathbb{R}^{3 \times 512 \times 512}$, UNI spatial tokens $\mathbf{U} \in \mathbb{R}^{32 \times 32 \times 1024}$ (Section~\ref{sec:uni_features}), and a conditioning token $y$, the model produces $\hat{\mathbf{x}}_{\text{IHC}} = G(\mathbf{x}_{\text{HE}}, \mathbf{U}, y)$.
The token $y$ encodes dataset-specific context: on BCI it represents the HER2 class ($y \in \{0, 1\!+, 2\!+, 3\!+\}$); on MIST it represents the target stain type ($y \in \{\text{HER2}, \text{Ki67}, \text{ER}, \text{PR}\}$).
In both cases, $y$ indexes into the same learnable embedding table and is injected via FiLM modulation (Section~\ref{sec:generator}).

The architecture has four components (Fig.~\ref{fig:architecture}):
(1)~a UNI feature processor that builds multi-scale conditioning maps from patch tokens,
(2)~a multi-scale edge encoder for structural features from H\&E,
(3)~a SPADE-UNet generator conditioned on UNI features (spatially) and class labels (channel-wise), and
(4)~an unconditional PatchGAN discriminator.

\subsection{UNI Feature Extraction and Processing}
\label{sec:uni_features}

We extract spatial features from a frozen UNI~\cite{chen2024uni} ViT-L/16.
Feeding the full $512\!\times\!512$ image as a single input would produce only $14\!\times\!14$ tokens at coarse spatial resolution.
Instead, we divide the image into a $4\!\times\!4$ grid of sub-crops, pass each through UNI independently, and reassemble and average-pool the resulting tokens into a $32\!\times\!32$ grid ($N\!=\!1{,}024$ tokens of dimension $d\!=\!1{,}024$).

A lightweight processor $\mathcal{P}$ adapts these tokens for the decoder:
\begin{equation}
    \mathbf{U}^{(s)} = \mathcal{P}_s(\mathbf{U}), \quad s \in \{32, 64, 128, 256\}
\end{equation}
Each token is projected from $d\!=\!1{,}024$ to $512$ dimensions and refined with residual convolutions at $32\!\times\!32$.
Learned transposed convolutions then produce resolution-matched maps for the remaining decoder stages.

\subsection{SPADE-UNet Generator}
\label{sec:generator}

The generator follows a UNet encoder-decoder with SPADE~\cite{park2019spade} conditioning.
Five strided convolution blocks downsample H\&E from $512$ to $16\!\times\!16$ ($3 \to 64 \to 128 \to 256 \to 512 \to 512$).
The bottleneck applies residual blocks with self-attention for global context.

Each decoder block receives upsampled features, the encoder skip connection, and edge features.
After channel reduction, a SPADE block applies dual conditioning:
\begin{equation}
    \mathbf{h}' = (\boldsymbol{\gamma}_{\text{UNI}} + \boldsymbol{\gamma}_{\text{cls}}) \odot \hat{\mathbf{h}} + (\boldsymbol{\beta}_{\text{UNI}} + \boldsymbol{\beta}_{\text{cls}})
    \label{eq:spade}
\end{equation}
where $\hat{\mathbf{h}} = \text{IN}(\mathbf{h})$ is instance-normalized.
\emph{Spatial modulation} $\boldsymbol{\gamma}_{\text{UNI}}, \boldsymbol{\beta}_{\text{UNI}} = \text{Conv}(\text{LeakyReLU}(\text{Conv}(\mathbf{U}^{(s)})))$ provides spatially-varying scale and bias from tissue morphology.
\emph{Channel modulation} $\boldsymbol{\gamma}_{\text{cls}}, \boldsymbol{\beta}_{\text{cls}} = \text{FiLM}(\mathbf{e}_y)$ from a learned conditioning embedding $\mathbf{e}_y \in \mathbb{R}^{64}$ controls overall staining identity and intensity (HER2 class on BCI, stain type on MIST).
SPADE parameters are zero-initialized (ControlNet-style~\cite{zhang2023controlnet}), so spatial modulation has no effect at the start of training and is learned gradually; FiLM parameters are initialized as $\boldsymbol{\gamma}_{\text{cls}}\!=\!1, \boldsymbol{\beta}_{\text{cls}}\!=\!0$, providing identity channel modulation at initialization.

\paragraph{Edge encoder.}
\label{sec:edge_encoder}
A lightweight multi-scale edge encoder concatenates H\&E RGB with Sobel gradient maps and extracts structural features at five scales ($512, 256, 128, 64, 32$) via two-layer CNNs.
Edge features are concatenated with the corresponding skip connections, providing pixel-aligned structural priors.
The final decoder output is concatenated with the original H\&E input before the output convolution, forming an input skip connection that eases color mapping.

\subsection{Misalignment-Aware Loss Design}
\label{sec:losses}

Both BCI~\cite{liu2022bci} and MIST~\cite{li2023asp} pair H\&E and IHC from consecutive tissue sections, introducing spatial misalignment between training pairs.
However, because the generator produces output directly from the H\&E input, the generated image and the input H\&E are inherently pixel-aligned.
We exploit this asymmetry in loss design.

\paragraph{Reconstruction (generated vs.\ real IHC).}
All paired losses operate at resolutions where misalignment is negligible:
\begin{equation}
    \mathcal{L}_{\text{percept}} = \text{LPIPS}_{128}(\hat{\mathbf{x}}, \mathbf{x}_{\text{IHC}}) + 0.5 \cdot \text{LPIPS}_{256}(\hat{\mathbf{x}}, \mathbf{x}_{\text{IHC}})
\end{equation}
\begin{equation}
    \mathcal{L}_{\text{L1}} = \| \hat{\mathbf{x}}^{(64)} - \mathbf{x}_{\text{IHC}}^{(64)} \|_1
\end{equation}

\paragraph{Structure preservation (generated vs.\ H\&E).}
Since the output and H\&E are pixel-aligned:
\begin{equation}
    \mathcal{L}_{\text{edge}} = \sum_{s \in \{512, 256\}} \| \nabla_{\text{Sobel}}(\hat{\mathbf{x}}^{(s)}) - \nabla_{\text{Sobel}}(\mathbf{x}_{\text{HE}}^{(s)}) \|_1
\end{equation}

\paragraph{Adversarial.}
We use a hinge adversarial loss with an \emph{unconditional} discriminator and $R_1$ gradient penalty, since a conditional discriminator on misaligned pairs learns spatial inconsistency as ``real.''
Feature matching provides misalignment-robust texture supervision:
\begin{equation}
    \mathcal{L}_{\text{FM}} = \sum_{l} \| D^{(l)}(\hat{\mathbf{x}}) - D^{(l)}(\mathbf{x}_{\text{IHC}}) \|_1
\end{equation}

\paragraph{DAB stain losses.}
DAB intensity is extracted via Beer--Lambert color deconvolution~\cite{ruifrok2001quantification}.
$\mathcal{L}_{\text{DAB}}$ matches the mean top-10\% DAB intensity per image.

\paragraph{Total generator loss.}
\begin{equation}
\begin{aligned}
    \mathcal{L}_G = &\;\mathcal{L}_{\text{percept}} + \lambda_{\text{L1}} \mathcal{L}_{\text{L1}} + \lambda_{\text{edge}} \mathcal{L}_{\text{edge}} \\
    &+ \mathcal{L}_{\text{adv}} + \lambda_{\text{FM}} \mathcal{L}_{\text{FM}} \\
    &+ \lambda_{\text{DAB}} \mathcal{L}_{\text{DAB}}
\end{aligned}
\end{equation}
with $\lambda_{\text{L1}}\!=\!1.0$, $\lambda_{\text{edge}}\!=\!0.5$, $\lambda_{\text{FM}}\!=\!10.0$, $\lambda_{\text{DAB}}\!=\!0.2$.

\subsection{Training Details}
\label{sec:training}

We use TTUR~\cite{heusel2017fid} with a 1{,}000-step linear warmup and delay adversarial training until step 2{,}000 to stabilize early generator learning.
During training, conditioning signals are randomly dropped (class labels 10\%, UNI features 10\%, both 5\%) to prevent the generator from relying on any single signal.
We use EMA (decay 0.999) for the generator at inference.
Full hyperparameters are in the supplementary material (Section~K).

\section{Experiments}
\label{sec:results}

\subsection{Experimental Setup}

\paragraph{Datasets.}
We evaluate on two public consecutive-section benchmarks (Table~\ref{tab:datasets}).
Both provide images at $1024\!\times\!1024$ native resolution; following prior work, we train on $512\!\times\!512$ random crops and evaluate on deterministic crops.
BCI provides WSI-level HER2 scores; MIST provides no HER2 grading labels, so the stain embedding encodes only the target marker identity.

\begin{table}[t]
\centering
\caption{Dataset summary. Both datasets pair H\&E and IHC from consecutive tissue sections.}
\label{tab:datasets}
\begin{tabular}{l l r r}
\toprule
Dataset & Stain & Train & Test \\
\midrule
BCI~\cite{liu2022bci} & HER2 & 3{,}896 & 977 \\
\midrule
\multirow{4}{*}{MIST~\cite{li2023asp}} & HER2 & 4{,}642 & 1{,}000 \\
 & Ki67 & 4{,}361 & 1{,}000 \\
 & ER & 4{,}153 & 1{,}000 \\
 & PR & 4{,}139 & 1{,}000 \\
\bottomrule
\end{tabular}
\end{table}

\paragraph{Metrics.}
We evaluate image quality with FID~\cite{heusel2017fid}, KID$\times$1k~\cite{binkowski2018kid}, and LPIPS~\cite{zhang2018lpips}.
SSIM~\cite{wang2004ssim} is included for comparability with prior work, but pixel-level metrics are unreliable under spatial misalignment~\cite{klockner2025h}.
For stain accuracy, we report Pearson-$r$ between per-image DAB intensity scores (mean of top-10\% pixels after color deconvolution~\cite{ruifrok2001quantification}) and DAB KL divergence (mean per-pair KL between 256-bin histograms, following~\cite{li2023odagan}).
Additional clinical metrics are in the supplementary material.

\paragraph{Compared methods.}
We compare against
ASP~\cite{li2023asp},
ODA-GAN~\cite{li2023odagan},
SIM-GAN~\cite{guan2025supervised},
PASB~\cite{pasb},
and USI-GAN~\cite{peng2025usigan}.
TDKStain~\cite{peng2024advancing} is excluded as it does not report on these datasets; TA-GAN~\cite{jiang2026topology} uses resized $256\!\times\!256$ images rather than the standard $512\!\times\!512$ protocol; PSPStain~\cite{liu2024pspstain} is excluded from the MIST comparison (too few reported metrics) but included in BCI.
All numbers are taken from the original published works; SIM-GAN trains on $512\!\times\!512$ resized images rather than crops.
We emphasize that reported values for the same method on the same dataset can vary substantially across papers: ASP~\cite{li2023asp} reports FID\,51.4 on MIST HER2, whereas re-evaluation by~\cite{li2023odagan} yields 89.3 and by~\cite{klockner2025h} yields 177, a ${>}3\times$ spread from differences in crop selection, preprocessing, and test splits.
Stain accuracy metrics are even less standardized: related works use different formulations of optical density correlation (e.g.\@ raw IoD difference vs.\ per-pixel Pearson-$r$ vs.\ integrated density correlation), making cross-paper comparison unreliable.
We therefore use image quality metrics (FID, KID, SSIM) for cross-method comparison, and reserve stain accuracy metrics (Pearson-$r$, DAB KL) for internal evaluations (ablation, resolution scaling) where all numbers come from the same pipeline.

All MIST results use a single unified model serving four stains; BCI uses a dedicated HER2 model with identical architecture.
We present results on MIST (Section~\ref{sec:mist}), BCI (Section~\ref{sec:bci}), unified vs.\ specialist comparison (Section~\ref{sec:unified}), and higher resolution (Section~\ref{sec:1024}).

% ─────────────────────────────────────────────────────────────────
%  4.2 Multi-Stain Results — MIST
% ─────────────────────────────────────────────────────────────────
\subsection{Multi-Stain Results: MIST}
\label{sec:mist}

Table~\ref{tab:mist_results} compares \ours{} against five recent methods spanning contrastive, domain-alignment, and optimal-transport objectives.
\ours{} obtains the best FID and KID on all four stains from a single unified model, whereas compared methods typically train a separate model per stain.
Per-image stain accuracy is consistent across all stains (Pearson-$r > 0.92$, DAB KL $< 0.19$).
A tissue-type stratified failure analysis (Section~\ref{sec:failure}) shows that remaining errors concentrate in non-tumor tissue, with $<$\,3\% failure rate on invasive carcinoma.
Per-stain specialist comparison is given in the supplementary material (Section~B).

\begin{table*}[t]
\centering
\caption{Quantitative comparison on MIST (1{,}000 test images per stain).
\ours{} uses a single unified model serving all four IHC markers via learned stain embeddings; compared methods typically train a separate model per stain.
All numbers are taken from the original papers.
Cells marked --- indicate the metric or stain was not reported.
\first{Best} and \secondbest{second best} highlighted per column.}
\label{tab:mist_results}
\resizebox{\textwidth}{!}{%
\begin{tabular}{l ccc ccc ccc ccc}
\toprule
& \multicolumn{3}{c}{HER2} & \multicolumn{3}{c}{Ki67} & \multicolumn{3}{c}{ER} & \multicolumn{3}{c}{PR} \\
\cmidrule(lr){2-4} \cmidrule(lr){5-7} \cmidrule(lr){8-10} \cmidrule(lr){11-13}
Method & FID$\downarrow$ & KID$\times$1k$\downarrow$ & SSIM$\uparrow$ & FID$\downarrow$ & KID$\times$1k$\downarrow$ & SSIM$\uparrow$ & FID$\downarrow$ & KID$\times$1k$\downarrow$ & SSIM$\uparrow$ & FID$\downarrow$ & KID$\times$1k$\downarrow$ & SSIM$\uparrow$ \\
\midrule
ODA-GAN~\cite{li2023odagan}      & 68.0 & \secondbest{8.6} & 0.189 & --- & --- & --- & --- & --- & --- & --- & --- & --- \\
PASB~\cite{pasb}                 & 46.0 & 21.0 & \first{0.274} & --- & --- & --- & 40.5 & 21.6 & \secondbest{0.254} & --- & --- & --- \\
SIM-GAN~\cite{guan2025supervised} & 39.6 & --- & 0.154 & 31.6 & --- & 0.182 & 34.6 & --- & 0.176 & 36.1 & --- & 0.183 \\
ASP~\cite{li2023asp}             & 51.4 & 12.4 & 0.200 & 51.0 & \secondbest{19.1} & \secondbest{0.241} & 41.4 & \secondbest{5.8} & 0.206 & 44.8 & \secondbest{10.2} & \secondbest{0.216} \\
USI-GAN~\cite{peng2025usigan}    & \secondbest{37.8} & --- & 0.187 & \secondbest{27.4} & --- & 0.232 & \secondbest{33.1} & --- & 0.202 & \secondbest{34.6} & --- & \secondbest{0.216} \\
\midrule
\ours{} (ours)                   & \first{34.5} & \first{2.2} & \secondbest{0.229} & \first{27.2} & \first{1.8} & \first{0.282} & \first{29.2} & \first{1.8} & \first{0.258} & \first{29.0} & \first{1.1} & \first{0.269} \\
\bottomrule
\end{tabular}
}
\end{table*}

\begin{figure}[t]
\centering
\includegraphics[width=\columnwidth]{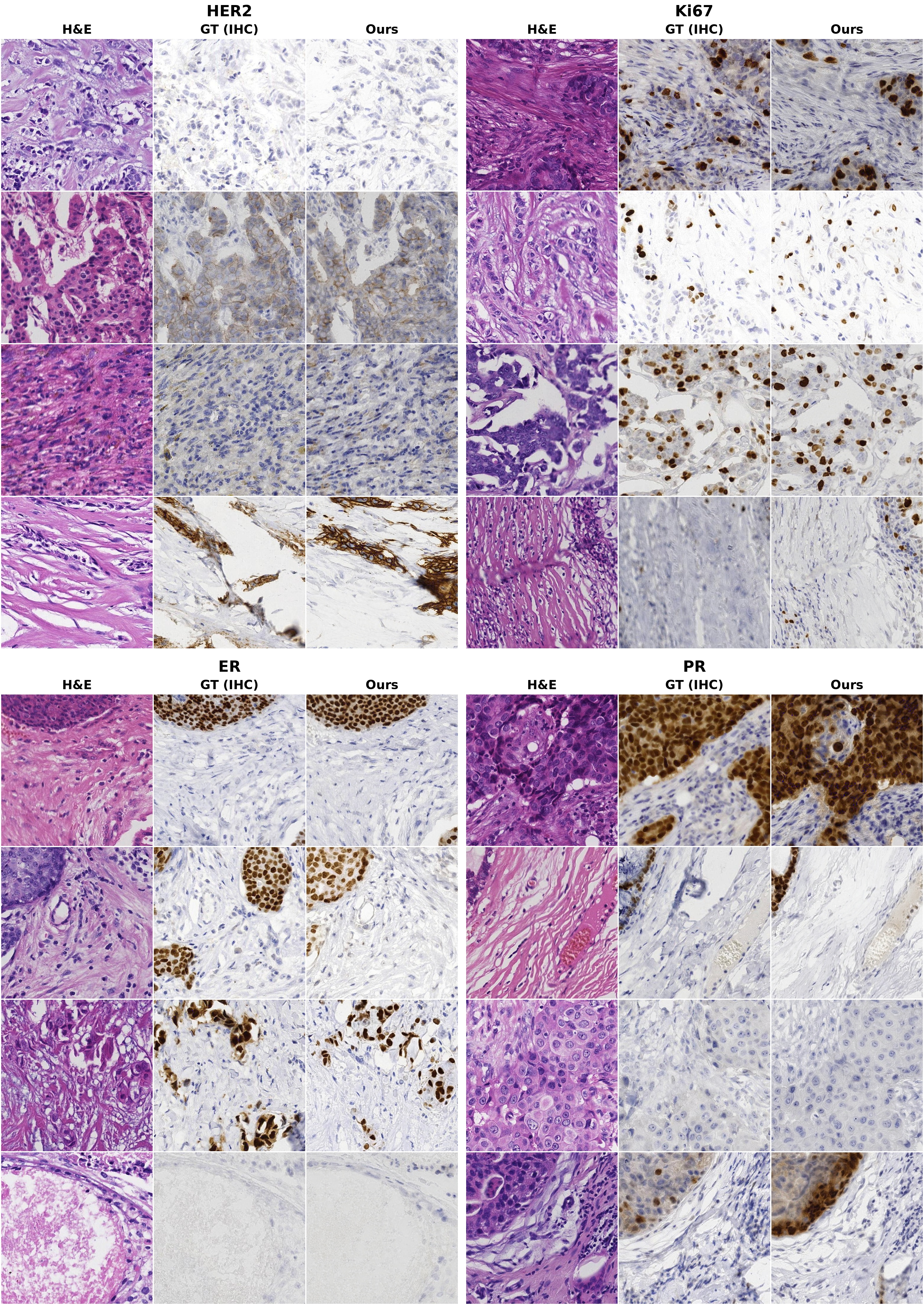}
\caption{Unified multi-stain generation on MIST. Four randomly sampled validation images per stain (HER2, Ki67, ER, PR). Columns: H\&E input, ground truth IHC, and \ours{} output. A single model produces stain-specific expression patterns: membrane (HER2), punctate nuclear (Ki67), and diffuse nuclear (ER/PR).}
\label{fig:mist_multistain}
\end{figure}

\begin{figure}[t]
\centering
\includegraphics[width=\columnwidth]{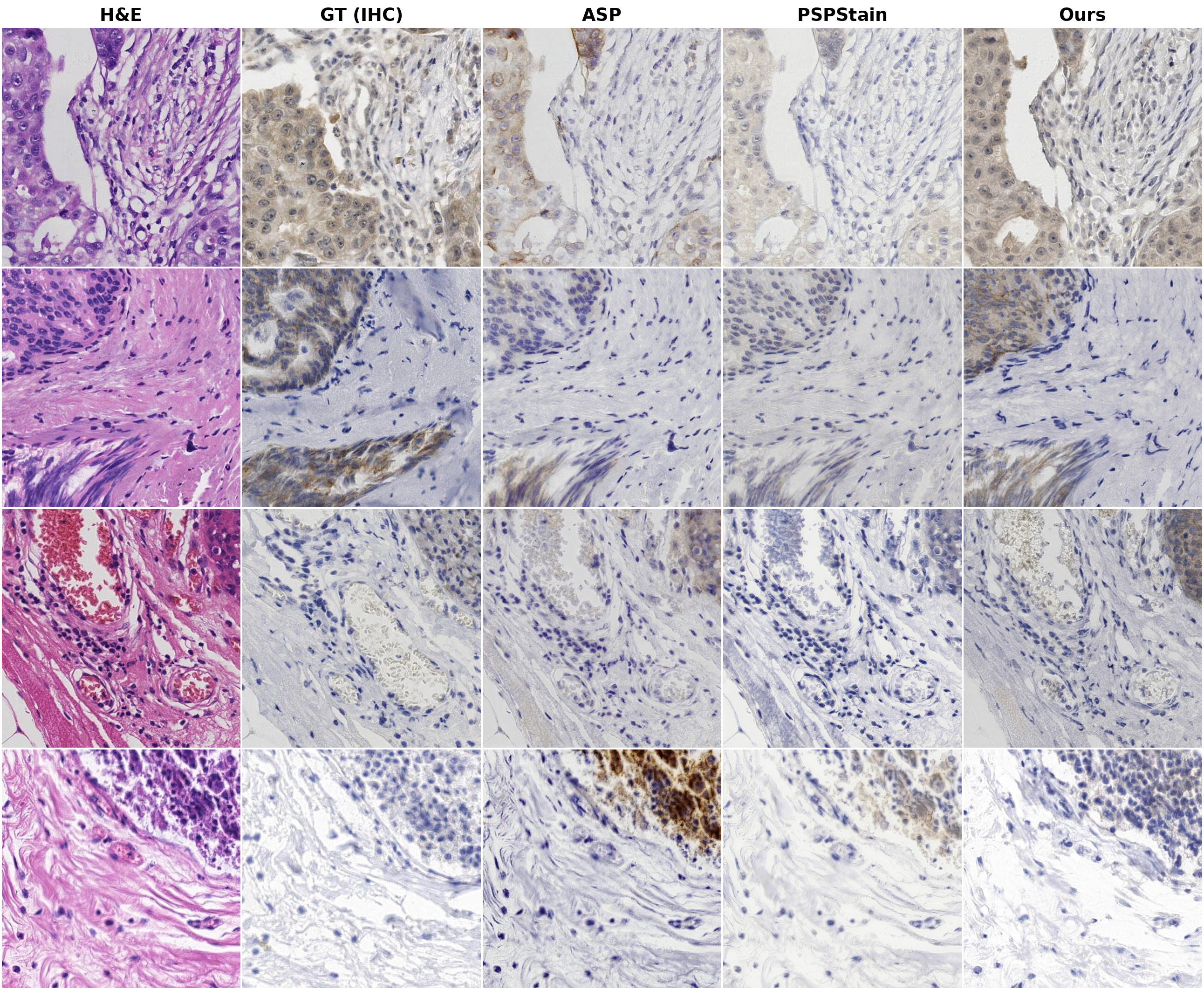}
\caption{Representative qualitative comparison on MIST HER2. Additional randomly sampled comparisons are provided in the supplementary material (Section~D).}
\label{fig:mist_comparison}
\end{figure}

Figure~\ref{fig:mist_multistain} illustrates the unified model's outputs across all four stains: membrane-associated DAB for HER2, punctate nuclear dots for Ki67, and diffuse nuclear staining for ER/PR, all from a single set of weights.
Figure~\ref{fig:mist_comparison} provides a side-by-side comparison on HER2.

% ─────────────────────────────────────────────────────────────────
%  4.3 Single-Stain Results — BCI
% ─────────────────────────────────────────────────────────────────
\subsection{Single-Stain Results: BCI}
\label{sec:bci}

We next validate on BCI, a dataset with different tissue sources and known spatial misalignment from consecutive sections.
\ours{} leads on FID, KID, and SSIM among all compared methods (Table~\ref{tab:bci_results}).
Stain accuracy is also strong: DAB KL of 0.482 vs.\ 2.933 for ODA-GAN, the only other method to report this metric on BCI, while Pearson-$r$ reaches 0.867.
Per-class DAB quantification confirms correct HER2 grading ($0 < 1\!+ < 2\!+ < 3\!+$); details are provided in the supplementary material.
Figure~\ref{fig:bci_comparison} shows representative outputs across all four HER2 classes.

\begin{table}[t]
\centering
\caption{Quantitative comparison on BCI test set (977 images).
All numbers are taken from the original papers.
$^\dagger$Evaluated on a 500-image test split.
$^\ddagger$As reported by~\cite{pasb}.
\first{Best} and \secondbest{second best} highlighted.}
\label{tab:bci_results}
\begin{tabular}{l c c c}
\toprule
Method & FID $\downarrow$ & KID$\times$1k $\downarrow$ & SSIM $\uparrow$ \\
\midrule
ASP$^\dagger$~\cite{li2023asp}       & 65.1  & 9.9   & \secondbest{0.503} \\
ODA-GAN$^\dagger$~\cite{li2023odagan} & 59.3  & 10.1  & 0.473 \\
PSPStain$^\ddagger$~\cite{liu2024pspstain} & 45.7  & 30.4  & 0.433 \\
PASB~\cite{pasb}                      & \secondbest{43.6}  & \secondbest{9.6}   & 0.426 \\
\midrule
\ours{} (ours)                        & \first{34.6}  & \first{6.5}  & \first{0.541} \\
\bottomrule
\end{tabular}
\end{table}

\begin{figure}[t]
\centering
\includegraphics[width=\columnwidth]{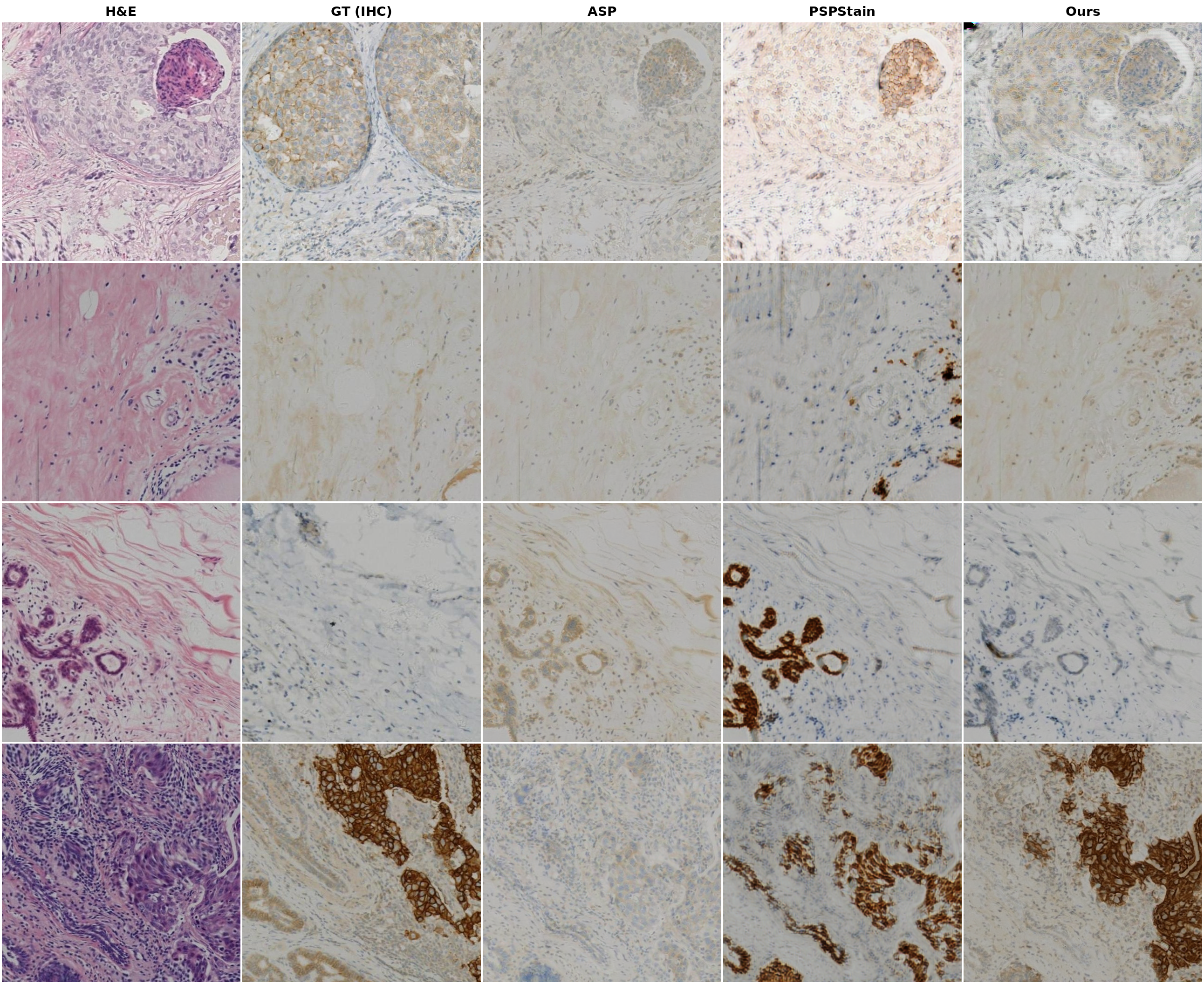}
\caption{Qualitative comparison on BCI (one sample per HER2 class). Columns: H\&E input, ground truth IHC, ASP~\cite{li2023asp}, PSPStain~\cite{liu2024pspstain}, and \ours{}. Our method preserves tissue morphology and produces correctly graded DAB staining across all HER2 levels.}
\label{fig:bci_comparison}
\end{figure}

% ─────────────────────────────────────────────────────────────────
%  4.4 Unified Multi-Stain Model
% ─────────────────────────────────────────────────────────────────
\subsection{Unified vs.\ Per-Stain Specialists}
\label{sec:unified}

A natural question is whether the unified model incurs a cost relative to training separate per-stain specialists.
Table~\ref{tab:unified} compares the unified model (42M trainable parameters, excluding the frozen UNI backbone) against four independently trained specialists (170M trainable total).

\begin{table}[t]
\centering
\caption{Unified model vs.\ per-stain specialists on MIST (macro-averaged across HER2, Ki67, ER, PR). The unified model matches specialist-level stain accuracy with $4\times$ fewer total parameters. Per-stain breakdown in supplementary material (Section~B).}
\label{tab:unified}
\resizebox{\columnwidth}{!}{%
\begin{tabular}{l c c c c c}
\toprule
Model & \#Models & Train.\ Params & Avg FID $\downarrow$ & Avg P-$r$ $\uparrow$ & Avg DAB KL $\downarrow$ \\
\midrule
Per-stain specialists & 4 & 170M & \first{29.8} & 0.930 & 0.171 \\
Unified (dim\,64)     & 1 & 42M  & 30.0         & \first{0.937} & \first{0.159} \\
\bottomrule
\end{tabular}
}
\end{table}

The unified model achieves comparable FID (30.0 vs.\ 29.8) while reducing total parameters by $4\times$.
Stain accuracy is comparable: the unified model slightly outperforms specialists on Pearson-$r$ (0.937 vs.\ 0.930) and DAB KL (0.159 vs.\ 0.171), though the differences are within typical run variance.
We also tested a larger 256-dimensional stain embedding; results were comparable (Avg FID 30.4, Avg P-$r$ 0.937), indicating that the model is insensitive to embedding dimensionality and a compact 64-d representation suffices.
Whether the learned stain embeddings capture biologically meaningful distinctions, not just aggregate statistics, is examined in the supplementary material (cross-stain generation in Section~E, stain fingerprint analysis in Section~F).

% ─────────────────────────────────────────────────────────────────
%  4.5 Scalability to 1024
% ─────────────────────────────────────────────────────────────────
\subsection{Scalability to Higher Resolution}
\label{sec:1024}

As a final test of architectural flexibility, we scale to native $1024\!\times\!1024$ generation by adding a single encoder/decoder level, increasing parameters by only 0.2\% (42.56M $\to$ 42.65M).
Table~\ref{tab:1024} compares $512$ and $1024$ models on both datasets.
On MIST, the $1024$ model improves stain accuracy (Pearson-$r$\,$0.961$ vs.\ $0.937$, DAB KL\,$0.099$ vs.\ $0.159$); FID is higher ($40.3$ vs.\ $30.0$), consistent with FID sensitivity to resolution and sample size.
On BCI, $1024$ generation matches $512$ image quality (FID $34.1$ vs.\ $34.6$) while improving stain accuracy (DAB KL $0.392$ vs.\ $0.482$).
Per-stain metrics are in the supplementary material (Section~C).

\begin{table}[t]
\centering
\caption{$512$ vs.\ $1024$ generation. MIST values are macro-averaged across 4 stains (unified model). $1024$ improves stain accuracy on both datasets with minimal parameter overhead.}
\label{tab:1024}
\begin{tabular}{l l c c c}
\toprule
Dataset & Res.\ & FID $\downarrow$ & P-$r$ $\uparrow$ & DAB KL $\downarrow$ \\
\midrule
\multirow{2}{*}{MIST} & $512$  & \first{30.0} & 0.937 & 0.159 \\
                       & $1024$ & 40.3         & \first{0.961} & \first{0.099} \\
\midrule
\multirow{2}{*}{BCI}  & $512$  & 34.6         & 0.867 & 0.482 \\
                       & $1024$ & \first{34.1} & \first{0.892} & \first{0.392} \\
\bottomrule
\end{tabular}
\end{table}

% ─────────────────────────────────────────────────────────────────
%  4.6 Failure Characterization
% ─────────────────────────────────────────────────────────────────
\subsection{Failure Characterization}
\label{sec:failure}

While aggregate metrics indicate strong overall performance, they can mask systematic failure modes.
To understand \emph{where} our model fails, we classify each 512$\times$512 H\&E test crop into seven breast tissue categories using CONCH~\cite{lu2024avisionlanguage} zero-shot classification (whole-image downsized to 224$\times$224) and measure per-image DAB KL divergence across all 4{,}000 MIST test images and 977 BCI test images.

\begin{figure}[t]
\centering
\includegraphics[width=\columnwidth]{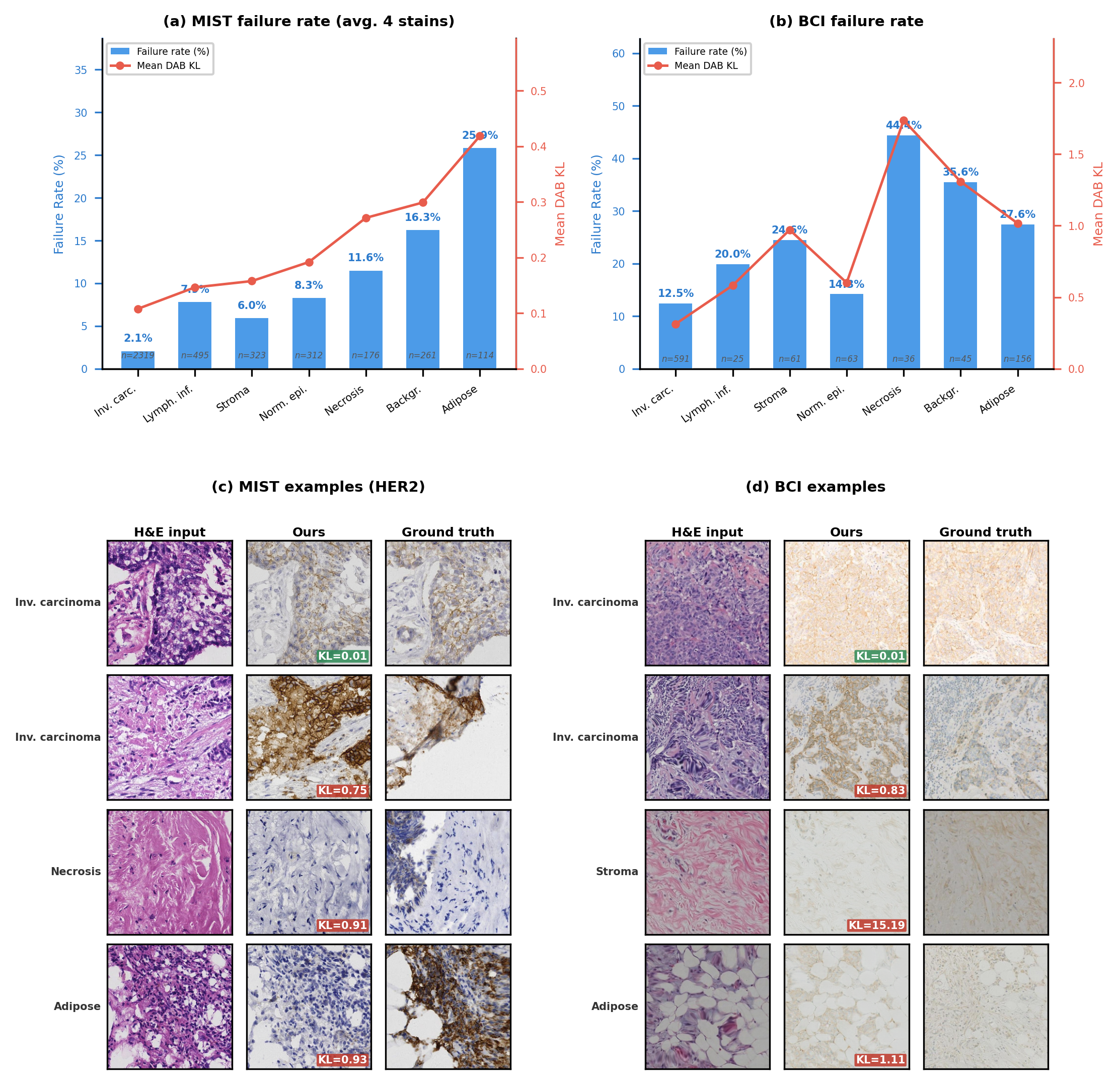}
\caption{Tissue-type stratified failure analysis via CONCH~\cite{lu2024avisionlanguage} zero-shot classification. (a,b)~Failure rate and mean DAB KL by tissue type on MIST (4{,}000 images, 4 stains, macro-averaged) and BCI (977 images). (c,d)~Representative examples: invasive carcinoma produces lower error, while adipose and necrotic regions produce higher error.}
\label{fig:failure_analysis}
\end{figure}

Figure~\ref{fig:failure_analysis}(a,b) stratifies our failure rate (DAB KL~$> 0.5$) by tissue type on MIST (macro-averaged across four stains) and BCI.
On invasive carcinoma, the most clinically relevant and prevalent tissue, only 2.1\% of MIST crops and 12.5\% of BCI crops exceed the failure threshold.
Failure rates rise for non-tumor tissues: adipose (25.9\% MIST, 27.6\% BCI), necrosis (11.6\% MIST, 44.4\% BCI), and background regions.
BCI failure rates are higher overall due to stronger consecutive-section misalignment, but the tissue-type ranking is broadly consistent across both datasets.
Failures concentrate in non-tumor tissue.

\begin{figure}[t]
\centering
\includegraphics[width=\columnwidth,height=0.95\textheight,keepaspectratio]{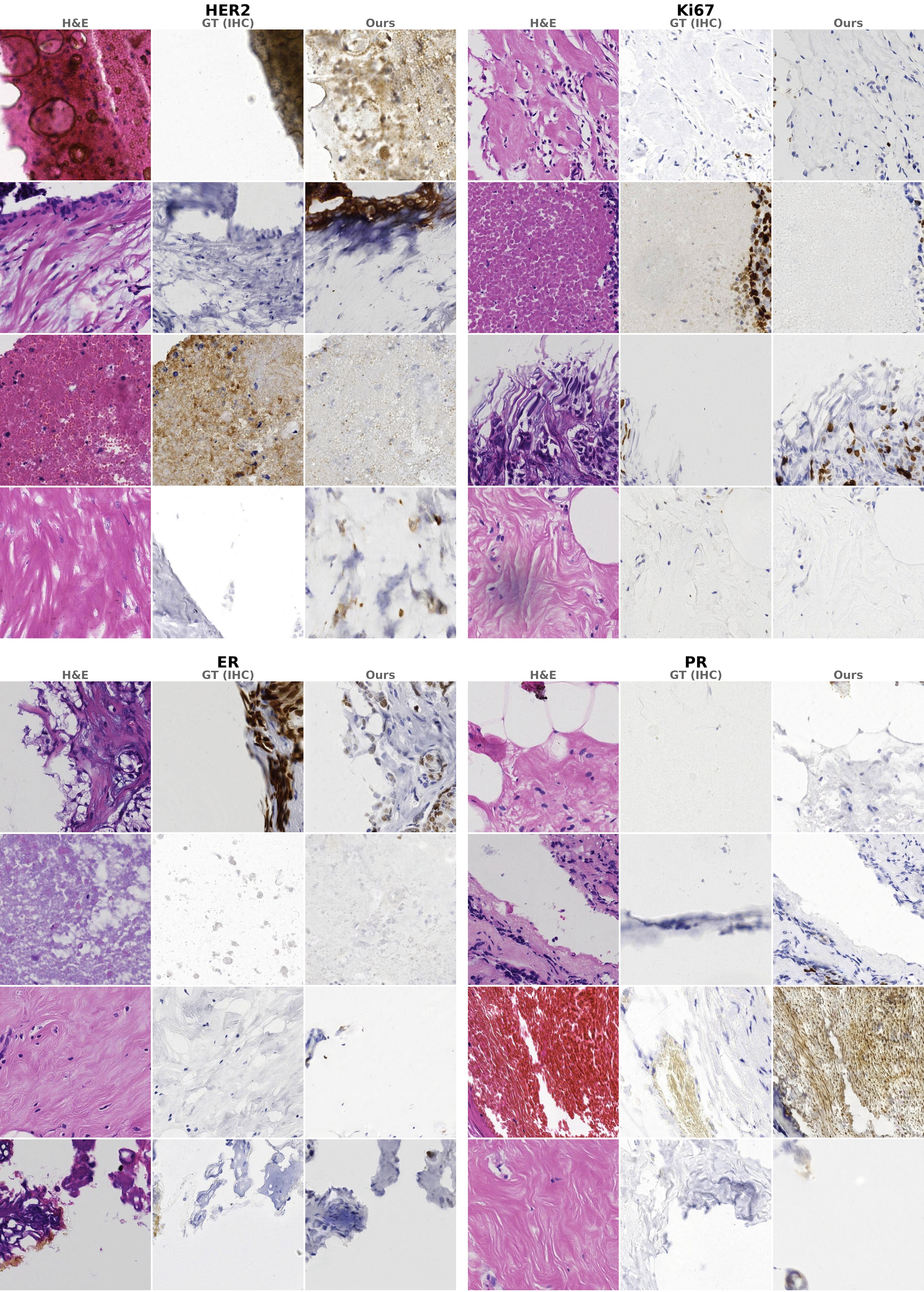}
\caption{Worst failure cases on MIST per stain, ranked by whole-image DAB KL divergence (top-4 per stain). Most failures occur in non-tumor tissue and where consecutive-section misalignment is more pronounced.}
\label{fig:worst_failures}
\end{figure}

Figure~\ref{fig:failure_analysis}(c,d) illustrates representative cases from each dataset.
A logistic classifier trained on frozen UNI CLS embeddings can predict failures across stains with AUC 0.72--0.85 (supplementary material, Section~J), consistent with failures being tissue-dependent rather than random.
In a deployment setting, such a predictor could serve as a preliminary quality flag to identify crops that may need manual review, though the moderate AUC suggests further development would be needed for reliable filtering.
Figure~\ref{fig:worst_failures} shows the four worst cases per stain on MIST; the corresponding BCI worst-case analysis is in the supplementary material (Section~J).

% ─────────────────────────────────────────────────────────────────
%  4.7 Ablation Study
% ─────────────────────────────────────────────────────────────────
\subsection{Ablation Study}
\label{sec:ablation}

We ablate key components using the unified 4-stain model on MIST (Table~\ref{tab:ablation}).
All variants share the same SPADE-UNet backbone, training schedule (100 epochs), and $512\!\times\!512$ crop protocol; only the ablated component is removed.
BCI ablations are reported in the supplementary material (Section~I).

\begin{table}[t]
\centering
\caption{Ablation study on MIST (macro-averaged across 4 stains, unified model). \first{Best} and \secondbest{second best} highlighted.}
\label{tab:ablation}
\begin{tabular}{l c c c}
\toprule
Config & FID $\downarrow$ & P-$r$ $\uparrow$ & DAB KL $\downarrow$ \\
\midrule
Full model          & \secondbest{30.0} & 0.937          & \secondbest{0.159} \\
$-$ Edge encoder    & 31.7          & \first{0.939}  & 0.162 \\
$-$ Feat.\ matching & \first{29.9}  & 0.932          & 0.171 \\
$-$ DAB loss        & 30.4          & 0.927          & 0.184 \\
$-$ LPIPS           & 31.4          & 0.931          & 0.189 \\
CLS $4\!\times\!4$  & 33.7          & \secondbest{0.938}  & \first{0.150} \\
\cmidrule(lr){1-4}
$-$ Discriminator   & 160.5         & 0.927          & 2.884 \\
$-$ UNI features    & 40.1          & 0.681          & 0.669 \\
\bottomrule
\end{tabular}
\end{table}

We organize findings along three axes: the UNI conditioning signal, architectural choices, and loss components.

Removing all spatial conditioning degrades every metric (FID $30.0 \to 40.1$, Pearson-$r$\,$0.937 \to 0.681$, DAB KL $0.159 \to 0.669$), confirming that dense spatial tokens are essential.
Reducing UNI spatial resolution from $32\!\times\!32$ to $4\!\times\!4$ (CLS $4\!\times\!4$) raises FID to 33.7 while Pearson-$r$ is essentially unchanged, indicating that dense spatial tokens improve image quality more than stain intensity on $512\!\times\!512$ crops.
Replacing UNI with a general-purpose DINOv3~\cite{simeoni2025dinov3} backbone degrades stain accuracy (DAB KL $+$79\%), confirming that pathology-specific pretraining matters; the full comparison is in supplementary Section~H.

For architectural choices, the discriminator is essential for image quality: without it, FID rises to 160.5 and DAB KL to 2.884, yet Pearson-$r$ drops only modestly ($0.937 \to 0.927$).
Reconstruction losses alone learn the correct mean staining intensity, but the discriminator is responsible for realistic texture and distribution-level accuracy.
The edge encoder is dataset-dependent: beneficial on MIST (FID 31.7 without it vs.\ 30.0), but negligible on BCI where consecutive-section misalignment limits edge reliability.

Among individual loss components, LPIPS contributes moderately: removing it raises FID to 31.4 and degrades stain accuracy (Pearson-$r$\,$0.937 \to 0.931$, DAB KL $0.159 \to 0.189$).
Feature matching presents a trade-off: FID is marginally better without it (29.9 vs.\ 30.0), but stain accuracy degrades (Pearson-$r$\,$0.932$, DAB KL $0.171$); it regularizes texture statistics toward realistic stain distributions at a slight cost to distributional diversity.
The DAB intensity loss has a targeted effect on stain accuracy: DAB KL rises from 0.159 to 0.184 and Pearson-$r$ drops from 0.937 to 0.927, while FID is essentially unchanged.

%% NOTE: tex/5_* appears to be missing from your original file.
%% If a section was accidentally omitted, add it here before tex/6_conclusion.
\section{Conclusion}
\label{sec:conclusion}

We have shown that conditioning a virtual staining generator on dense spatial tokens from a frozen pathology foundation model improves both image realism and stain quantification accuracy.
Three design choices drive the effectiveness of \ours{}:
(i)~$32\!\times\!32$ UNI spatial tokens provide tissue-level semantic guidance at each decoder stage;
(ii)~a misalignment-aware loss suite combining multi-scale LPIPS, an unconditional discriminator, and DAB intensity supervision produces accurate stain quantification despite the spatial shifts inherent in consecutive-section training pairs;
and (iii)~FiLM-based stain conditioning enables a single model to serve four IHC markers, matching per-stain specialists on stain accuracy at $4\times$ fewer total parameters.

On MIST, the unified model achieves the best distributional metrics across all four stains; on BCI, it also achieves the best distributional metrics.
A tissue-type stratified failure analysis shows that remaining errors are systematic and concentrate in non-tumor tissue.

The architecture scales to native $1024\!\times\!1024$ with only 0.2\% additional parameters; on BCI, $1024$ generation matches $512$ image quality (FID 34.1 vs.\ 34.6) while improving stain accuracy, and on MIST, stain accuracy improves further (Pearson-$r$\,$0.961$, DAB KL\,$0.099$).
All experiments use DAB-based IHC stains from two breast cancer datasets; generalization to other chromogens, tissue types, scanners, and study sites has not been tested, and as noted by~\cite{klockner2025h}, cross-site robustness remains largely unexplored in virtual staining.
Whether these remaining gaps reflect data limitations or fundamental ambiguity in the H\&E-to-IHC mapping, characterizing them systematically can help guide future efforts.
Extending the stain embedding to unseen markers through few-shot conditioning and validating clinical utility through downstream diagnostic tasks such as automated HER2 scoring are natural next steps.

%% ── Appendix / Supplementary ────────────────────────────────────
%% For arXiv, appending supplementary inline is strongly recommended.
%% Uncomment the block below if you have a supplementary tex file.
%%
%% \appendix
%% \input{tex/supplementary}

\bibliographystyle{ACM-Reference-Format}
\bibliography{references}

\end{document}